\documentclass{article}

\usepackage[final]{corl_2025} 

\usepackage{times} 
\usepackage{microtype}
\usepackage{hyperref}
\usepackage{xspace}
\usepackage{xcolor}
\definecolor{linkcolor}{RGB}{0,62,116}
\definecolor{webblue}{RGB}{25,118,210}
\hypersetup{
    colorlinks=true,
    urlcolor=webblue
}

\usepackage{graphicx}
\usepackage{subfigure}
\usepackage{booktabs}

\usepackage{bm}
\usepackage{amssymb}
\usepackage{mathtools}
\usepackage{amsthm}
\usepackage{cleveref}

\theoremstyle{plain}
\newtheorem{theorem}{Theorem}[section]

\theoremstyle{definition}
\newtheorem{definition}[theorem]{Definition}

\theoremstyle{remark}

\usepackage{soul}
\definecolor{lightgray}{RGB}{230, 230, 230}
\DeclareRobustCommand{\hlgray}[1]{{\sethlcolor{lightgray}\hl{#1}}}
\usepackage[textsize=tiny]{todonotes}
\usepackage{setspace}
\usepackage{footnote}

\usepackage{algorithm}
\usepackage{algorithmic}
\usepackage{enumitem}

\renewcommand{\vec}[1]{{\boldsymbol{#1}}}
\newcommand{\mat}[1]{{\boldsymbol{#1}}}

\newcommand{\ursa}{URSA\xspace}
\newcommand{\ours}{\ursa}
\newcommand{\oursLong}{Unsupervised Real-world Skill Acquisition\xspace}
\newcommand{\estimatedEntropy}{\widehat{\entropySymbol}}
\newcommand{\sizeRepertoire}{N_{\repertoire}}
\newcommand{\numDistinctSkills}{N_\skill}
\newcommand{\skillSpaceReachable}{\mathcal{Z}_{p}}
\newcommand{\skillSpaceAchieved}{\mathcal{Z}_{\policy}}
\newcommand{\volume}[1]{\mathrm{vol}\left(#1\right)}
\newcommand{\surrogateProb}{q}
\newcommand{\diff}{\mathrm{d}}
\newcommand{\distanceFunc}{d}
\newcommand{\lse}{\mathrm{LSE}}
\newcommand{\dimSkillSpace}{D}
\newcommand{\periodSampleSkills}{T}
\newcommand{\constant}{k}

\newcommand{\oursITE}{\ours$+$ ITE\xspace}

\newcommand{\cost}{c}
\newcommand{\costCritic}{C}

\newcommand{\repertoire}{\mathcal{R}}

\newcommand{\kde}[1]{\mathrm{\textsc{kde}}\left( {#1} \right)} 

\newcommand{\state}{\vec s}
\newcommand{\stateSpace}{\mathcal{S}}
\newcommand{\stateSpaceSafe}{\mathcal{S}_{\mathrm{safe}}}

\newcommand{\action}{\vec a}
\newcommand{\actionSpace}{\mathcal{A}}

\newcommand{\reward}{r}

\newcommand{\feat}{\vec \phi}

\newcommand{\featFunc}{\phi}

\newcommand{\discount}{\gamma}

\newcommand{\skill}{\vec z}

\newcommand{\skillImg}{\widetilde{\skill}}
\newcommand{\skillEnv}{\skill}
\newcommand{\skillSpace}{\mathcal{Z}}

\newcommand{\horizonImg}{H}
\newcommand{\replay}{\mathcal{D}}

\newcommand{\params}{\theta}

\newcommand{\policy}{\pi}
\newcommand{\policySkill}{\policy_\skill}

\newcommand{\lagrange}{\lambda}
\newcommand{\valueFunction}{V}

\newcommand{\successorFeatures}{\vec \psi}
\newcommand{\costFunction}{C}

\newcommand{\wm}{\mathcal{W}}
\newcommand{\stateImg}{\widetilde{s}}

\newcommand{\uniform}{\mathcal{U}}
\newcommand{\normal}{\mathcal{N}}

\newcommand{\expect}[2]{\mathbb{E}_{{#1}} \left[{#2}\right]}

\newcommand{\norm}[1]{\left\lVert#1\right\rVert_2}
\newcommand{\threshold}{\delta}

\newcommand{\entropySymbol}{\mathrm{H}}
\newcommand{\entropy}[1]{\entropySymbol\left( {#1} \right)}

\newcommand{\qdac}{QDAC\xspace}

\newcommand{\qdacLong}{Quality-Diversity Actor-Critic\xspace}

\newcommand{\daydreamer}{DayDreamer\xspace}
\newcommand{\dominic}{DOMiNiC\xspace}

\makeatletter
\newcommand{\subalign}[1]{%
  \vcenter{%
    \Let@ \restore@math@cr \default@tag
    \baselineskip\fontdimen10 \scriptfont\tw@
    \advance\baselineskip\fontdimen12 \scriptfont\tw@
    \lineskip\thr@@\fontdimen8 \scriptfont\thr@@
    \lineskiplimit\lineskip
    \ialign{\hfil$\m@th\scriptstyle##$&$\m@th\scriptstyle{}##$\hfil\crcr
      #1\crcr
    }%
  }%
}
\makeatother

\makeatletter
\AfterEndEnvironment{algorithm}{\let\@algcomment\relax}
\AtEndEnvironment{algorithm}{\kern2pt\hrule\relax\vskip3pt\@algcomment}
\let\@algcomment\relax
\newcommand\algcomment[1]{\def\@algcomment{\footnotesize#1}}

\renewcommand\fs@ruled{\def\@fs@cfont{\bfseries}\let\@fs@capt\floatc@ruled
  \def\@fs@pre{\hrule height.8pt depth0pt \kern2pt}%
  \def\@fs@post{}%
  \def\@fs@mid{\kern2pt\hrule\kern2pt}%
  \let\@fs@iftopcapt\iftrue}
\makeatother

\definecolor{perf}{HTML}{E1144B}
\definecolor{dist}{HTML}{0053D6}
\definecolor{safe}{HTML}{b28704}

\newif\ifcomments
\commentstrue 

\setlist{leftmargin=5.5mm}
\setboolean{ALC@noend}{true}

\title{From Tabula Rasa to Emergent Abilities:\\Discovering Robot Skills via\\Real-World Unsupervised Quality-Diversity}

\author{Luca~Grillotti \hspace{2em}Lisa~Coiffard \hspace{2em}Oscar~Pang \hspace{2em}Maxence~Faldor \hspace{2em}Antoine~Cully\\
  Adaptive \& Intelligent Robotics Lab\\
  Imperial College London
}

\usepackage{bbm}

\DeclareMathOperator*{\argmin}{arg\,min}

\begin{document}

\maketitle

\begin{figure}[h]
    \centering
    \includegraphics[width=\textwidth]{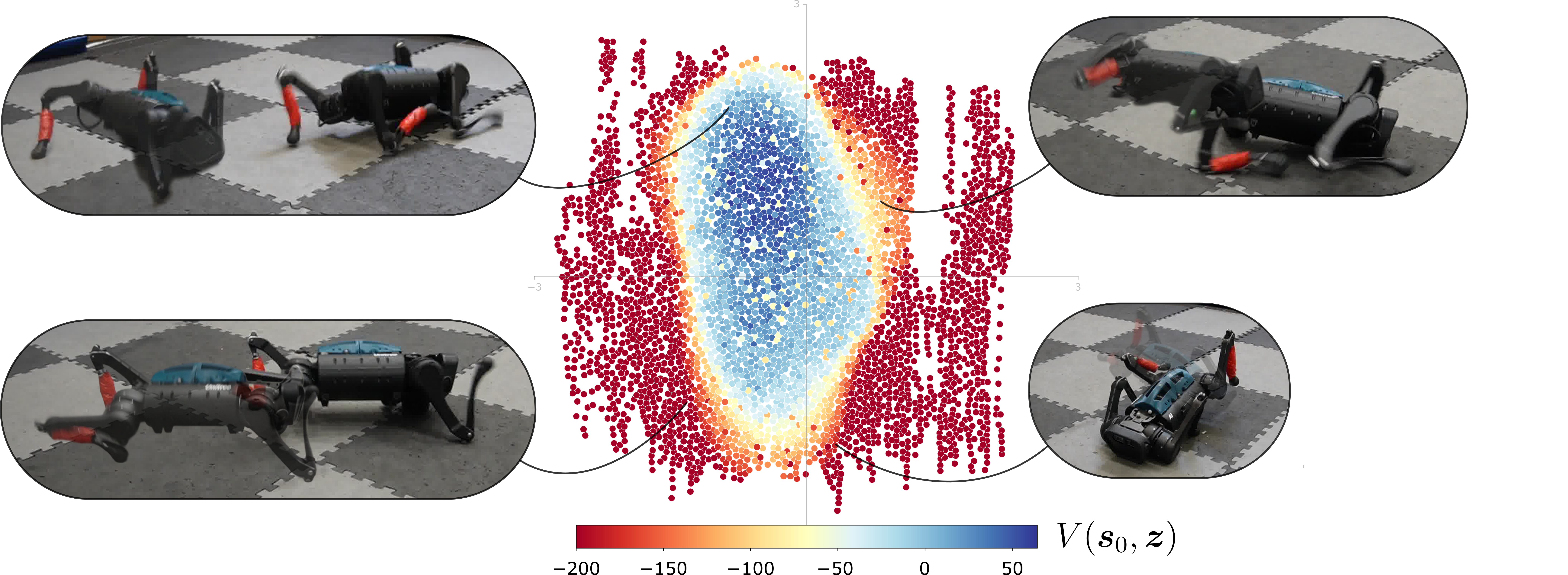}
    \vspace{-1em}
    \caption{We propose \oursLong (\ours), a framework for unsupervised quality-diversity in real-world environments. Each skill is plotted in the latent space with color indicating its estimated value, $V(\state_0, \skill)$, highlighting the diversity of learned behaviors.}
    \label{fig:repertoire}
\end{figure}

\begin{abstract}
    Autonomous skill discovery aims to enable robots to acquire diverse behaviors without explicit supervision. Learning such behaviors directly on physical hardware remains challenging due to safety and data efficiency constraints. Existing methods, including Quality-Diversity Actor-Critic (\qdac), require manually defined skill spaces and carefully tuned heuristics, limiting real-world applicability. We propose \oursLong (\ours), an extension of \qdac that enables robots to autonomously discover and master diverse, high-performing skills directly in the real world. We demonstrate that \ours successfully discovers diverse locomotion skills on a Unitree A1 quadruped in both simulation and the real world. Our approach supports both heuristic-driven skill discovery and fully unsupervised settings. We also show that the learned skill repertoire can be reused for downstream tasks such as real-world damage adaptation, where \ours outperforms all baselines in 5 out of 9 simulated and 3 out of 5 real-world damage scenarios. Our results establish a new framework for real-world robot learning that enables continuous skill discovery with limited human intervention, representing a significant step toward more autonomous and adaptable robotic systems.
    Demonstration videos are available at \href{https://adaptive-intelligent-robotics.github.io/URSA/}{adaptive-intelligent-robotics.github.io/URSA}.
\end{abstract}

\section{Introduction}

Animals exhibit the remarkable ability to develop diverse movement patterns through autonomous interaction with their environment. Inspired by this, autonomous skill discovery aims to endow robots with a similar capacity: the ability to learn a wide repertoire of behaviors through self-directed experience. Such behavioral diversity is critical for general-purpose robots that must operate in dynamic, unpredictable environments. While specialized robots have shown strong performance in domains like PCB insertion~\citep{Luo2024SERLAS} and legged locomotion~\citep{WalkInThePark}, they remain limited to a narrow set of pre-defined behaviors. In contrast, a robot equipped with a diverse skill set can adapt to new tasks and conditions, such as recovering from damage~\citep{cully_RobotsThatCan_2015, chatzilygeroudis2018reset, kaushik2020adaptive} and navigating on planar terrains \citep{dadsoff, resetfreeQDRealworld}.

Quality-Diversity (QD) algorithms~\citep{pugh2016quality, cully2017quality} offer a promising foundation for building such skill repertoires, with the aim to discover a set of behaviors that are both diverse (e.g. different gaits) and individually high-performing (e.g. forward velocity). While traditional QD methods rely on manually defined behavior descriptors, recent work~\citep{aurora, ruda, aurora-xcon} learns these skill representations directly from data, in an unsupervised manner. However, applying QD to physical robots remains challenging—either it depends on simulation, which requires accurate modelling and reliable sim-to-real transfer, or it must operate directly in the real world, where methods like evolutionary algorithms tend to be too sample-inefficient for practical deployment.

In this work, we introduce \oursLong{} (\ours), a framework for unsupervised QD that operates directly on physical robots. \ours{} extends the \qdacLong{} (\qdac) framework \citep{airl2024qdac} for unsupervised learning and leverages the \daydreamer{}~\citep{daydreamer} world model as its backbone, enabling efficient skill acquisition directly on hardware without relying on simulation or sim-to-real transfer. We evaluate \ours{} on a Unitree A1 quadruped and show it outperforms baselines in 5 out of 9 simulated and 3 out of 5 real-world damage scenarios. This demonstrates its potential for enabling adaptable, resilient skill discovery directly on physical robots.
Our key contributions are:
\vspace{-0.8em}
\begin{itemize}
    \item We propose \ours{}, a framework for unsupervised QD that extends \qdac with a learned latent skill space, safety-aware optimization, and efficient sampling.
    \item We show that \ours{} can learn diverse locomotion behaviors directly in the real world using imagination-based training with world models.
    \item We demonstrate that the learned skill repertoire enables adaptation to physical damage by supporting selective skill reuse in downstream tasks.
\end{itemize}
\vspace{-0.8em}

\section{Problem Statement: A New Perspective on Skill Discovery}

We model the environment as a Markov Decision Process (MDP) $(\stateSpace, \actionSpace, r, p)$~\citep{sutton_ReinforcementLearningIntroduction_2018}, where at each timestep $t$, the agent in state $\state_t \in \stateSpace$ takes an action $\action_t \in \actionSpace$, transitions to $\state_{t+1} \sim p(\cdot | \state_t, \action_t)$, and receives reward $\reward_t = r(\state_t, \action_t)$. We assume access to a feature function $\featFunc: \stateSpace \times \actionSpace \to \skillSpace$, either learned or user-defined, that captures instantaneous behavioral properties, such as velocity or foot contact patterns in a robot. 

To summarize behavior over time, we define a \emph{skill} as the expected feature vector under a policy's stationary distribution. 
\begin{definition}[Skill]
    \label{def:skill}
    The skill $\skill \in \skillSpace$ of a policy $\policy$ is defined as $\skill = \expect{\policy}{\featFunc(\state, \action)}$.
\end{definition}

This expectation captures the characteristic behavior induced by a policy. For example, in a quadrupedal robot where $\feat_t[i] = 1$ if the $i$-th foot is on the ground and $0$ otherwise, the skill $\skill$ represents the episodic proportion of contact time per foot. A vector like $\skill = \begin{bmatrix}0.8 & 0.3 & 0.8 & 0.3\end{bmatrix}^\intercal$ describes a gait where the left feet are used 80\% of the time and the right feet 30\%, potentially corresponding to a limping motion.

Within the skill space $\skillSpace$, we distinguish two subspaces: the reachable skill space of all theoretically attainable skills, and the achieved skill space of skills actually mastered by our policy.

\begin{definition}[Reachable Skill Space]
    The reachable skill space $\skillSpaceReachable \subseteq \skillSpace$ is the set of all skills $\skill\in\skillSpace$ for which some policy $\policy$ can achieve them: $\skillSpaceReachable = \left\{ \skill \in \skillSpace \mid \exists \policy, \expect{\policy}{\featFunc(\state, \action)} = \skill \right\}$
\end{definition}

\begin{definition}[Achieved Skill Space of a Policy]
    A skill $\skill\in\skillSpace$ is said to be \textit{achieved} by $\policy$ if and only if $\expect{\policySkill}{\featFunc(\state, \action)} = \skill$, where $\policySkill$ represents the skill-conditioned policy $\policy(\cdot | \cdot, \skill)$.
    The achieved skill space $\skillSpaceAchieved \subseteq \skillSpace$ is the set of all achieved skills: $\skillSpaceAchieved = \left\{ \skill \in \skillSpace \mid \expect{\policySkill}{\featFunc(\state, \action)} = \skill \right\}$
\end{definition}

By construction, it follows that $\skillSpaceAchieved \subseteq \skillSpaceReachable \subseteq \skillSpace$. In this work, we aim to learn these spaces, with the ultimate goal of (1) maximizing the volume of the achieved skill space $\volume{\skillSpaceAchieved}$, thereby encouraging behavioral diversity, and (2) ensuring that each skill is performed with high expected return. However, jointly optimizing for both diversity and performance across a potentially high-dimensional skill space poses significant challenges.

\begin{figure*}[t]
    \centering
    \includegraphics[width=0.85\textwidth]{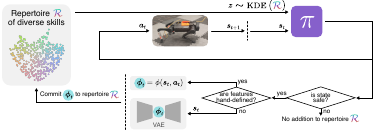}
    \vspace{-0.2cm}
    \caption{Overview of \ours: The system checks if the state $\state_t$ is safe, if so, encodes it into features $\feat_t$, and builds a diverse skill repertoire $\repertoire$. New skills are sampled using a Kernel Density Estimator on the repertoire from the safe, reachable skill space. The skill-conditioned policy $\policySkill$ maximizes its expected return while matching the sampled skill $\skill$.}
    \label{fig:ursa_diagram}
\end{figure*}

To make this problem tractable, we introduce a surrogate probability distribution $q(\cdot)$ over $\skillSpace$, designed to approximate a uniform distribution over $\skillSpaceReachable$ and decompose the objective into two components:
\vspace{-0.8em}
\begin{itemize}
    \item \textbf{Diversity:} Maximize the entropy of the skill distribution $q$ over the reachable skill space $\skillSpaceReachable$.
    \item \textbf{Performance:} Learn a skill-conditioned policy $\policySkill$ that maximizes the expected return while achieving each sampled skill $\skill \sim q$:
\end{itemize}
\vspace{-1em}
\begin{equation}
    \label{eq:problem-1}
    \text{maximize} \quad \expect{\policySkill}{\sum_{i=0}^{\infty} \discount^i \reward_{t+i}} \quad
    \text{subject to} \quad \expect{\policySkill}{\featFunc(\state, \action)} = \skill
    \tag{P1}
\end{equation}

\section{Background}

\subsection{Quality-Diversity Actor-Critic (QDAC)}

Our proposed method builds upon the QDAC algorithm~\citep{airl2024qdac}, and extends it to unsupervised QD in a real-world setting. QDAC aims at finding a skill-conditioned policy $\policySkill$ that maximizes the reward while following a given skill $\skill$, effectively tackling problem~\ref{eq:problem-1}. It does this by making two approximations: (1) the expected state is approximated by the discounted average $(1 - \discount) \successorFeatures(\state, \skill)$, and (2) the strict equality constraint from problem~\ref{eq:problem-1} is replaced by an inequality constraint forcing the policy to stay close to the target skill $\skill$. This defines a new problem:
\begin{align}
\label{eq:problem-2}
\forall \skill \sim \uniform\left(\skillSpaceReachable\right),\quad \text{maximize } {\color{perf} \valueFunction(\state, \skill)} \quad \text{subject to } {\color{dist} \norm{(1 - \discount) \successorFeatures(\state, \skill) - \skill}} \leq \threshold
\tag{P2}
\end{align}
where $\valueFunction(\state, \skill) = \expect{\policySkill}{\left.\sum_{i=0}^{\infty} \discount^i \reward_{t+i}\right\vert \state_t = \state}$ is the value function~\citep{sutton_ReinforcementLearningIntroduction_2018}, $\successorFeatures(\state, \skill) = \expect{\policySkill}{\left.\sum_{i=0}^{\infty} \discount^i \feat_{t+i}\right\vert \state_t = \state}$ successor features~\citep{barreto_SuccessorFeaturesTransfer_2017} and $\threshold$ is a hyperparameter that determines the maximal acceptable distance between the expected state and the skill.

Our approach addresses two core limitations of \qdac. First, rather than assuming a predefined reachable skill space $\skillSpaceReachable$ and a fixed constraint threshold $\threshold$, we learn both the structure of the skill space and how to efficiently sample from it during training, while adaptively tuning $\threshold$. Second, instead of relying on manually defined features, we extract them from unsupervised encodings of the agent's state, enabling skill discovery directly from raw observations.

\subsection{Real-world Robot Learning with DayDreamer}

DayDreamer~\citep{daydreamer} is a model-based RL algorithm that enables efficient real-world robot learning by predicting environment dynamics in a latent space using a recurrent state-space model (RSSM). Its asynchronous architecture comprises a learner thread for model and policy updates and an actor thread for data collection, allowing continuous training and effective use of limited real-world interactions through imagined rollouts. We build on this framework by integrating unsupervised QD into both threads, enabling robots to autonomously explore and master diverse behaviors.

\section{Unsupervised Real-world Skill Acquisition}

We propose \oursLong (\ursa), an RL framework for autonomously discovering diverse skills in real-world environments without prior knowledge of the skill space. Built on \qdac~\citep{airl2024qdac}, \ursa introduces three extensions: (1) safety constraints to prevent unsafe behaviors, (2) a skill repertoire $\repertoire$ to store skills, and (3) an optional learnable feature function $\featFunc(\cdot)$ for compact skill representations from observations. \ursa leverages the \daydreamer{} architecture~\citep{daydreamer}, running two asynchronous threads: one for collecting real-world data (Algorithm~\ref{algo:ursa_collect}), and another for updating the world model and policy via imagined rollouts (Algorithm~\ref{algo:ursa_train}, see Figure~\ref{fig:ursa_training_actor} in \cref{appendix:imagination-training}).

\subsection{Ensuring Safe Skill Discovery}

In real-world robotics, some skills, such as falling, can damage the robot and should be avoided. 
To handle this, we use a mechanism following the constrained RL~\citep{altman2021cmdp} framework.
More specifically, we define a user-provided cost function $\cost: \stateSpace \to \mathbb{R}$ such that $\cost(\state) \leq 0$ if and only if $\state$ is in a hand-defined set of safe states $\stateSpaceSafe$.
We ensure that the associated critic $\costCritic(\state, \skill)=\expect{\policySkill}{\sum_{i=0}^{\infty} \discount^i \cost_{t+i} | \state_t = \state}$ remains non-positive for every skill $\skill$ learned by \ours.
Incorporating safety into skill discovery yields the constrained optimization problem (see Appendix~\ref{appendix:lagrange-details} for further implementation details):
\begin{equation}
    \label{eq:problem-3}
    \text{maximize } {\color{perf} \valueFunction(\state, \skill)} \quad
    \text{subject to } {\color{dist} \norm{(1 - \discount) \successorFeatures(\state, \skill) - \skill}} \leq \threshold \quad
    \text{and } {\color{safe} \costCritic(\state, \skill)} \leq 0
    \tag{P3}
\end{equation}

\begin{figure*}[t]
    \begin{minipage}[t]{0.495\textwidth}
    \input{algorithms/ursa_collect}
    \end{minipage}
    \hfill
    \begin{minipage}[t]{0.495\textwidth}
    \input{algorithms/ursa_train}
    \end{minipage}
\end{figure*}

\subsection{Efficient Skill Sampling from the Reachable Space}
\label{sec:method-skill_sampling}

To maximize the diversity of achieved skills, \ours must efficiently explore the reachable skill space $\skillSpaceReachable$ while ensuring safe execution. \ours addresses this by maintaining a diversity-aware repertoire $\repertoire$ and sampling skills from a flexible surrogate distribution $\surrogateProb(\cdot)$ learned from experience. Figure~\ref{fig:ursa_diagram} and Algorithm~\ref{algo:ursa_collect} provide high-level overviews of the skill collection and sampling process.

\paragraph{Adaptive Skill Distribution via Non-Parametric Density Estimation}
We model $\surrogateProb$ as a Gaussian Kernel Density Estimator (KDE)~\citep{parzen1962kde, rosenblatt1956kde}, which provides a flexible, non-parametric way to capture the structure of the reachable skill space for sampling physically achievable skills. We fit this distribution on the repertoire $\repertoire=\{\skill_{i}\}_{i=1}^{\sizeRepertoire}$ as $\surrogateProb = \kde{\repertoire} = \frac{1}{\sizeRepertoire} \sum_{i=1}^{\sizeRepertoire} \normal(\skill_i, \mat{\Sigma})$, where $\mat{\Sigma}$ is the repertoire's skill covariance matrix, with a scaling factor of $\sizeRepertoire^{-\frac{1}{\dimSkillSpace+4}}$ (where $\dimSkillSpace$ denotes the dimensionality of the skill space), following the approach of~\citet{scott1992multivariate}.

\paragraph{Maximizing Entropy for Uniform Skill Sampling}
To approximate a uniform sampling of $\skillSpaceReachable$, we maximize the entropy of the sampling distribution $\surrogateProb$ by maintaining a diverse repertoire. This entropy can be approximated via Monte-Carlo sampling with the skills from the repertoire: $\entropy{\surrogateProb} =-\int \surrogateProb(x) \log \surrogateProb(x) \diff x \quad\approx\quad  -\frac{1}{\sizeRepertoire} \sum_{i=1}^{\sizeRepertoire} q(\skill_i) \log q(\skill_i)$. 
This estimated entropy, written $\estimatedEntropy(\surrogateProb)$, has a tractable lower bound based on nearest-neighbor Mahalanobis distances $\distanceFunc_{\mat{\Sigma}}$, derived in Appendix~\ref{appendix:math-derivations}:
\begin{equation}
    \label{eq:ineqEntropy}
    \estimatedEntropy(\surrogateProb) \geq - \frac{1}{\sizeRepertoire} \sum_{i=1}^{\sizeRepertoire} \log \left( 1 + (\sizeRepertoire - 1) e^{ -\frac{1}{2} \distanceFunc_{\mat{\Sigma}}(\skill_i, \skill_i^{\mathrm{nn}})^2 } \right) + \frac{1}{2} \log \left(\det(\mat{\Sigma}) \right) + \constant
\end{equation}
where $\constant$ is a constant and $\skill_i^{\mathrm{nn}}$ is the nearest-neighbor of skill $\skill_i$ in the repertoire. To maximize this bound, we continuously update the repertoire by replacing existing skills with newly discovered ones that increase the distribution's spread.

\paragraph{Filling the Repertoire with Safe and Reachable Skills}

We maximize diversity within the reachable space by adding features that result in \textit{safe} states to the repertoire. When a new feature $\feat_t$ is committed, we compute distances $\distanceFunc_{\mat{\Sigma}}(\skill, \skill^{\mathrm{nn}})$ for all $\skill \in \repertoire \cup \{\feat_t\}$, and remove the skill with the smallest distance, thereby keeping the repertoire size fixed. Assuming the impact of those updates on $\mat{\Sigma}$ is negligible, \ursa maximizes the lower bound on the approximated entropy of the skill-sampling distribution $\surrogateProb$ (see Equation~\ref{eq:ineqEntropy}). This process maintains a uniform coverage of the safe and reachable skill space, while continuously increasing diversity through the discovery of new skills.

\paragraph{Dynamic Threshold}

As the repertoire expands, the typical distance between skills increases, requiring adaptive tuning of the constraint threshold $\threshold$ in Problem~\ref{eq:problem-3}. To that end, we introduce a hyperparameter $\numDistinctSkills$, which controls the number of distinct skills that the robot can execute. We compute $\threshold$ as the mean nearest-neighbor distance, if the repertoire contained $\numDistinctSkills$ uniformly distributed skills (see equation in Appendix~\ref{appendix:adaptive-threshold}).

This ensures that executing one skill does not inadvertently satisfy multiple constraints.

\subsection{Unsupervised Quality-Diversity}

When a feature function is not provided, \ours learns one automatically from raw state observations. We implement $\featFunc(\cdot)$ using a variational autoencoder (VAE)~\cite{vae2013}, which encodes zeroth-order kinematics (e.g., joint angles, body height) from $\stateSpace$ into a compact latent representation. 

Following Definition~\ref{def:skill}, a skill $\skill$ is defined as the expected latent encoding under the policy's stationary distribution. The VAE is trained on the diverse set of states collected in the repertoire $\repertoire$, enabling unsupervised discovery of meaningful skills without human supervision.

\section{Results}
\label{sec:results}

\subsection{Experimental Setup and Baselines}

We evaluate \ours{} on the Unitree A1 quadruped in both PyBullet simulation and real-world environments. The robot state includes joint angles and velocities for its 12 joints; actions specify target joint positions executed via a 20\,Hz PD controller. Skills are sampled every 250 steps from a fixed repertoire of $N_R = 4096$. Training in simulation runs for 1 million timesteps, while in the real world, we collect 5 hours of data with asynchronous data collection and training on a single NVIDIA RTX 6000 Ada GPU. To match real-world conditions, episodes in simulation proceed continuously without manual resets to predetermined states. Additional training details and all hyperparameters are provided in Appendix~\ref{appendix:training-details}.
Demonstration videos are available at \href{https://adaptive-intelligent-robotics.github.io/URSA/}{adaptive-intelligent-robotics.github.io/URSA}.

We compare against two baselines: \textbf{\daydreamer}~\cite{daydreamer}, a single-skill learner corresponding to \ours{} without the successor features constraint in~\ref{eq:problem-3}; and \textbf{\dominic}~\cite{dominic}, which incorporates successor features and enforces a near-optimality constraint, discovering multiple policies only after achieving a high reward. Following the original implementation, we set the number of skills in \dominic to 8.
To ensure a fair comparison, we implemented \dominic using the same DayDreamer backbone as our method, maintaining architectural consistency across all compared approaches.

Our evaluation addresses three key questions: \textbf{(RQ1)} Can \ours learn diverse, high-performing and unsupervised locomotion skills? \textbf{(RQ2)} Can these skills be reused for downstream tasks like damage adaptation? \textbf{(RQ3)} Can \ours discover useful, controllable skills for target-driven tasks?

\begin{figure}[t]
    \centering
    \includegraphics[width=1\textwidth]{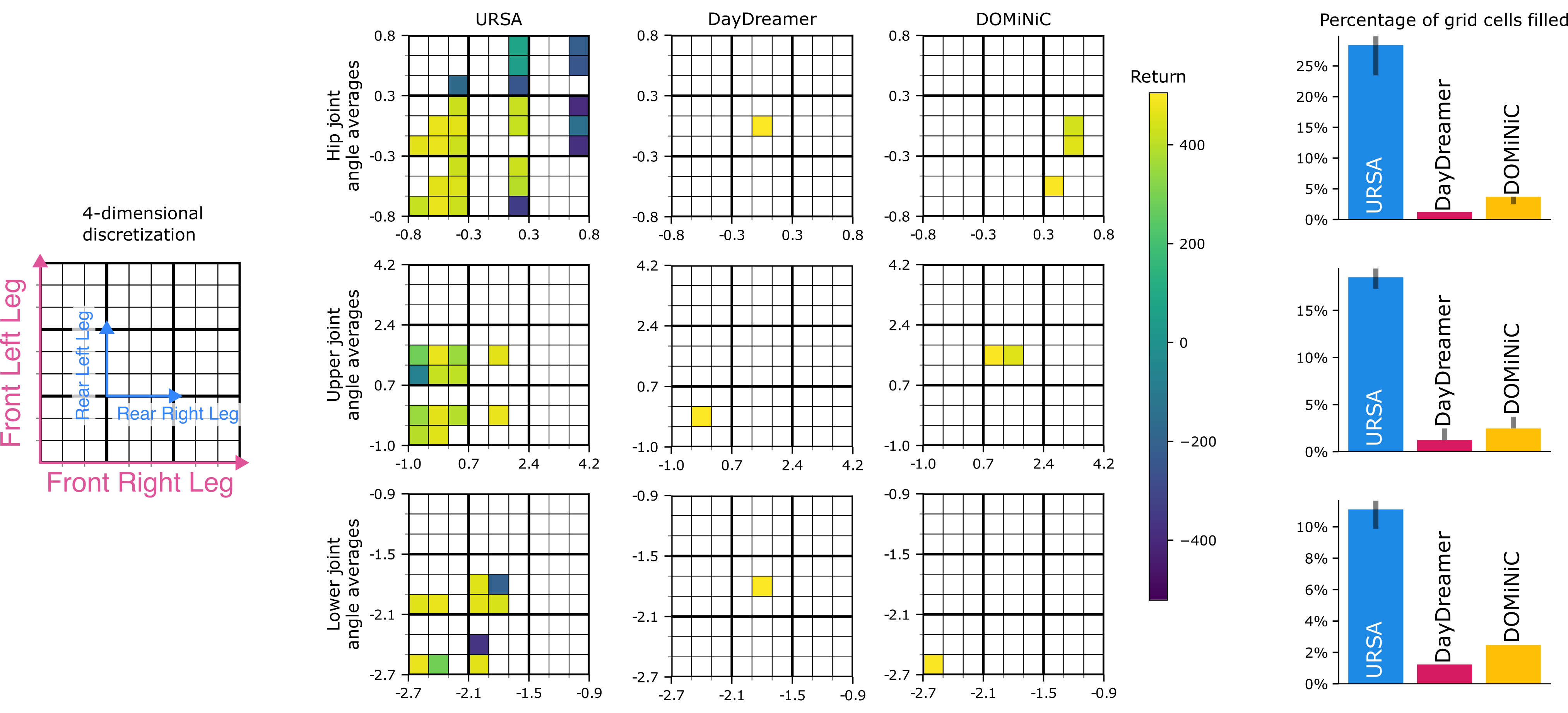}
    \vspace{-2em}
    \caption{Average joint angles across the skill repertoire in \ours, \daydreamer, and \dominic. Each cell represents average joint angles (hip, upper, and lower) for all leg combinations. Cells are colored if at least one skill’s average falls within that region.}
    \label{fig:angles_coverage}
\end{figure}

\subsection{RQ1: Unsupervised Discovery of Diverse and High-Performing Locomotion Skills}\label{sec:rq1}

To evaluate \ours's ability to learn diverse locomotion behaviors in an unsupervised setting, we train it to acquire forward movement skills without predefined behavioral objectives. Features $\feat_t$ are represented as 2-dimensional latent vectors encoded from raw states using a VAE. The reward function encourages forward velocity while maintaining stability, and the hand-defined set of safe states $\stateSpaceSafe$ ensures upright posture (see Appendix~\ref{appendix:reward-unsupervised} for details).

Figure~\ref{fig:angles_coverage} illustrates the distribution of average joint angles, evaluated in simulation, across different leg combinations. \ours discovers a wide variety of motion patterns through varied use of the leg joints. 
This is further supported by Figure~\ref{fig:repertoire}, which visualizes four example behaviors from the repertoire, ranging from gaits that keep the torso near the ground to more agile, coordinated walking motions. By contrast, \daydreamer{} converges on a single locomotion strategy that maximizes return, while \dominic{}—though able to learn multiple skills—enforces near-optimality, resulting in less joint-level diversity across its 8 discrete behaviors. 
Quantitatively, when projecting the joint angle trajectories onto discretized spaces of average angles, \ours achieves 4 times greater coverage compared to the baselines, demonstrating significantly more behavioral diversity (Fig.~\ref{fig:angles_coverage}).

\subsection{RQ2: Downstream Use of Skills for Damage Adaptation}
\label{sec:results-rq2}

\begin{figure}[t]
    \centering
    \includegraphics[width=\textwidth]{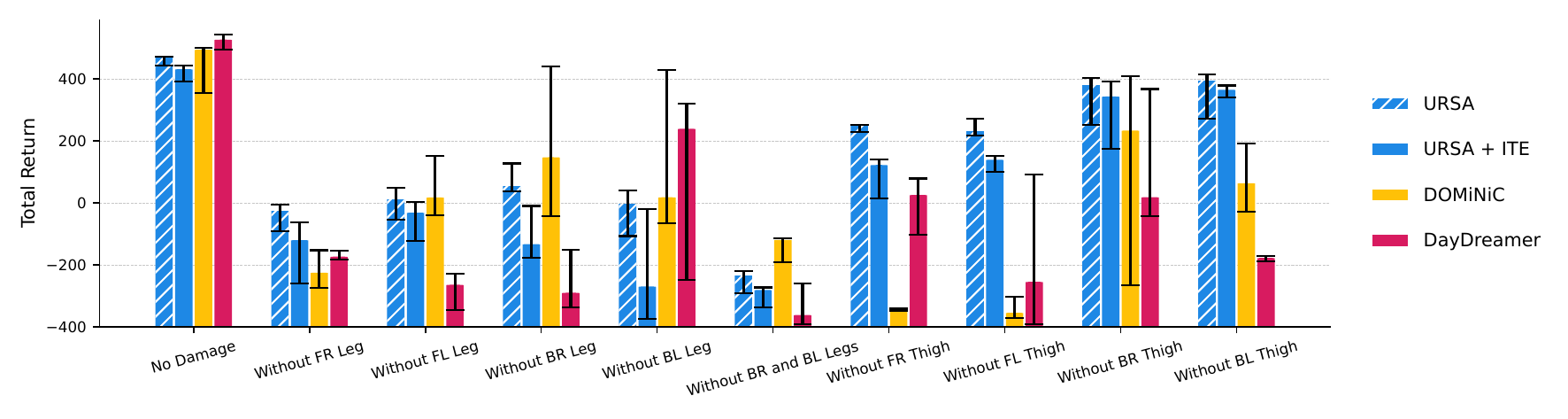}
    \vspace{-2em}
    \caption{Comparison of returns across joint damage scenarios in simulation. The best return for \ours is shown in hatched bars, compared to a version using ITE for adaptation. Results display the median return and interquartile range (IQR) across 5 independent runs.}
    \label{fig:damages}
\end{figure}

\begin{figure}
    \centering
    \includegraphics[width=1\linewidth]{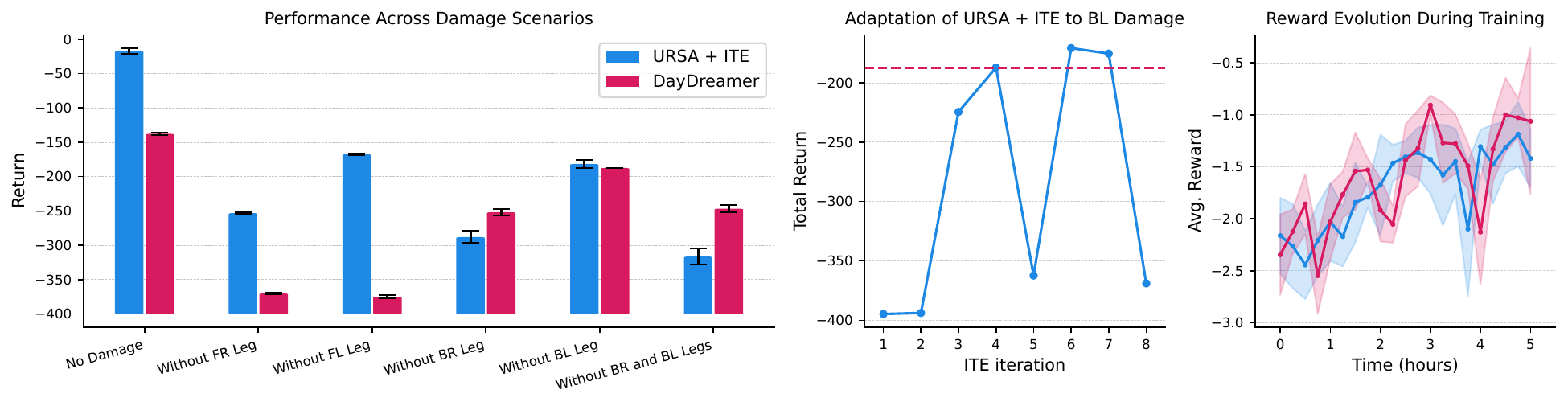}
    \vspace{-1.5em}
    \caption{\textbf{Left}: Comparison of returns across damage scenarios in the real world, showing median return and IQR across 2 independent runs. \textbf{Middle}: Evolution of attempted skills for a single run of ITE with FL leg damage. \textbf{Right}: Average reward during training (shaded area as standard deviation of 15-minute segments).}
    \label{fig:damages-rw}
\end{figure}

To evaluate the utility of \ours's learned repertoire for downstream tasks, we study its effectiveness in adapting to physical damage across simulation (Figure~\ref{fig:damages}) and real-world (Figure~\ref{fig:damages-rw}) scenarios. We apply damage scenarios including single joint failures and full failures of one or two legs\footnote{Joint notation: F = Front, B = Back, R = Right, L = Left (e.g., FR = Front Right, BL = Back Left).}.

In Figure~\ref{fig:damages}, we report an upper bound for \ours by executing all skills in the repertoire and recording the highest return. Given the large number of distinct skills (see $N_z$ in Appendix~\ref{appendix:hyperparameters}), we also evaluate \ours combined with Iterative Trial and Error (ITE)~\cite{cully_RobotsThatCan_2015} (\oursITE), which uses Bayesian optimization to find the best-performing skill in a few trials (see Appendix~\ref{appendix:ite}). \oursITE outperforms both \dominic and \daydreamer in upper-leg joint damage. However, with more severe damage (e.g., full leg failures), the near-optimality constraint of \dominic proves more beneficial than \ours's diversity-aware mechanism.

In Figure~\ref{fig:damages-rw}, we evaluate only \oursITE due to the impracticality of executing all skills. \ours outperforms \daydreamer in all damage scenarios except for back-right leg failures (left panel). Notably, \ours demonstrates reduced overall performance degradation under damage, leveraging its diverse skill repertoire to compensate for broken joints. This is evident in Figure~\ref{fig:damages-rw} (middle), where \oursITE progressively improves total return after consecutive iterations, recovering within 8 iterations during back-left leg damage. The underlying skill repertoire's quality is shown in Figure~\ref{fig:damages-rw} (right), where consistent improvement in average rewards indicates that \ours builds a diverse set of high-performing behaviors. In contrast, \daydreamer tends to overfit to forward locomotion strategies, leading to significant performance drops when the front legs are damaged. Additionally, \ours achieves higher returns in the undamaged setting, suggesting that behavioral diversity not only improves robustness but also acts as a stepping stone toward discovering high-performing locomotion strategies, as opposed to single-skill optimization.

\subsection{RQ3: Heuristic-Based Skill Discovery for Velocity Tracking}
To assess whether \ours{} can discover useful and controllable behaviors, we set $\featFunc(s_t) = [v_{x,t},\, \omega_{z,t}]$ to capture forward and angular velocities. The reward encourages stable posture and smooth movement, and a hand-designed safe state set maintains upright poses (see Appendix~\ref{appendix:reward-velocity}).

\begin{figure}[t]
    \centering
    \includegraphics[width=\textwidth]{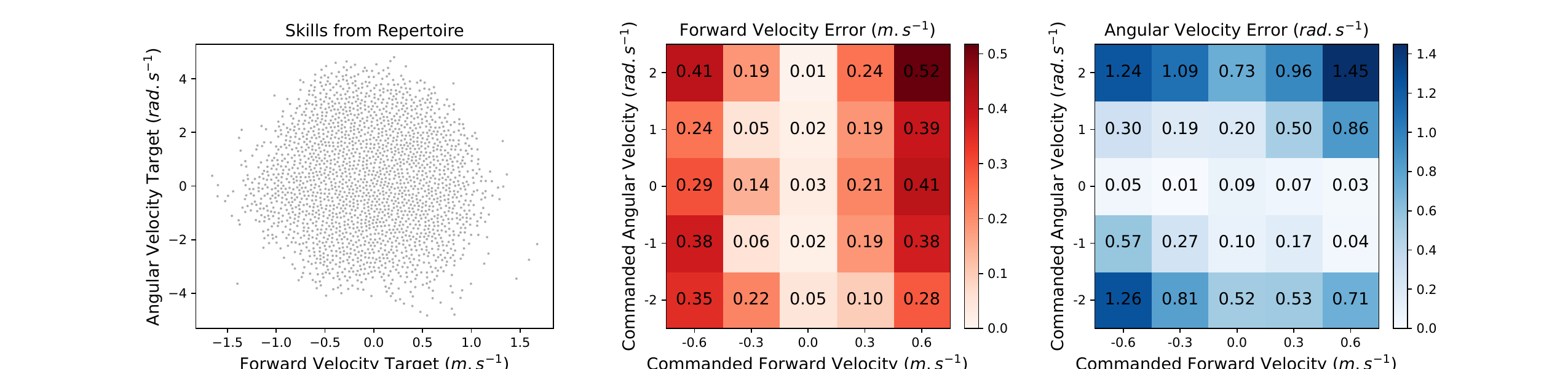}
    \vspace{-1.5em}
    \caption{Velocity tracking errors during skill execution, evaluating the robot's accuracy in following target velocity commands across the reachable space discovered by \ours. Lower values indicate better control.}
    \label{fig:velocities_error}
\end{figure}

We assess the usefulness of the discovered skills by measuring how well they track target velocity commands. Figure~\ref{fig:velocities_error} reports the tracking error across a grid of desired forward and angular velocities. \ours learns a repertoire that spans a broad range of these velocities, as confirmed by the plotted skill distribution (Fig.~\ref{fig:velocities_error}, left). However, we observe that tracking performance degrades as target velocities deviate further from zero, particularly in the case of forward velocity. This discrepancy likely arises because some target velocities, while instantaneously achievable, cannot be consistently sustained across the entire duration of a skill. These findings highlight that \ours is capable not only of discovering diverse behaviors, but also of learning skill representations that are suitable for real-world, target-driven control.

\section{Related Work}

\paragraph{Unsupervised skill discovery} Recent methods aim to discover diverse behaviors without explicit rewards. 
Unsupervised RL approaches \citep{diayn, dads, lsd, metra} aim at maximizing the mutual information between abstract skills and trajectories to induce distinct behaviors, while SMERL~\citep{smerl} and DOMiNO~\citep{domino} improve robustness by learning multiple solutions per task. In the Quality-Diversity (QD) domain, methods such as TAXONS~\citep{taxons}, STAX~\citep{stax}, and AURORA~\citep{aurora} learn behavioral descriptors directly from data, removing the need for hand-crafted behavioral descriptors. However, these approaches heavily rely on simulation and face challenges in real-world applications. Our work builds on these methods by introducing safety constraints and efficient skill sampling for real-world learning, as well as imagination-based training to reduce physical interactions.

\paragraph{Real-world robot learning} Real-world learning faces challenges like including limited data, safety concerns, and the lack of reset mechanisms. Reset-free QD~\citep{lim_LearningWalkAutonomously_2022, resetfreeQDRealworld} selects behaviors that self-reset. \citet{laversanne2021intrinsically} demonstrate diverse skill acquisition in robotic arms via goal-directed exploration, but still rely on episodic resets. off-DADS~\citep{dadsoff} enables skill discovery on real quadrupeds with off-policy RL, but does not consider any extrinsic reward. \daydreamer~\citep{daydreamer} sets a standard for sample-efficient learning using world models, and similarly, SERL~\citep{Luo2024SERLAS}, A Walk in the Park~\citep{WalkInThePark} and CrossQ~\citep{bohlinger2025gait} improve sample efficiency through careful system and algorithmic innovations. Our approach, like \daydreamer, leverages  world models for unsupervised multi-skill discovery, adding safety constraints, and sophisticated skill selection mechanisms.

\section{Discussion and Future Work}

In this work, we introduce URSA, showcasing its ability to perform safe and efficient unsupervised QD in the real world. We show that URSA successfully adapts to damage, leveraging its learned repertoire for resilience across diverse real-world damage scenarios. Although our results are promising, several challenges remain. One such challenge is the potential benefit of learning safety boundaries directly from data, rather than relying on predefined constraints. This approach could enhance the system's robustness in dynamic and changing environments. Furthermore, developing more advanced sampling strategies that focus on the agent's zone of proximal development—where the potential for learning progress is maximized—could improve the efficiency of skill acquisition. These advancements would represent steps toward more autonomous and adaptable robotic systems capable of continuously learning and evolving in real-world conditions.

\section{Limitations}

Although our approach demonstrates promising results, there are several important limitations to address. While URSA uses a skill-conditioned policy network that enables the discovery of diverse skills, some discovered skills naturally share similar gaits that rely on specific limbs, making them more vulnerable to targeted damage scenarios affecting those particular limbs. This architectural choice, while beneficial for training efficiency, can lead to performance degradation in specific damage conditions.

Additionally, URSA's performance is critically dependent on the selection of input state variables provided to the VAE, which requires careful engineering and domain knowledge. The choice of which state dimensions to include or exclude can significantly impact downstream task performance, particularly in damage adaptation scenarios. For instance, omitting rear leg joint information would likely result in less diverse rear leg behaviors and degraded performance when rear legs are damaged. However, predicting the optimal set of input variables remains challenging, as the relationship between VAE input features and the policy's capability to solve specific tasks is often difficult to establish~\citep{tarapore2016different}.

Our method also lacks an explicit mechanism, such as a dedicated reward function, to actively encourage the discovery of diverse skills. Instead, the system is limited to exploring skills that have been previously encountered during training, potentially missing novel and useful behaviors.

Moreover, our implementation is not fully reset-free. When the robot moves outside the designated arena, human intervention is still required to return it to the training area. This limitation could potentially be addressed by developing or providing a simple return-to-arena policy~\citep{minimalhumanintervention}, enabling the robot to autonomously reset its position when needed.

In tasks involving forward movement, our current reward function does not sufficiently penalize lateral and angular velocities. This results in behaviors that deviate from straight-line motion, indicating a need to refine the reward structure to better align with desired movement patterns.

Finally, while our system trains to maximize rewards across all discovered skills, it does not discriminate between high-potential skills, with effective forward movement, and less useful ones, with minimal movement (see Fig.~\ref{fig:repertoire}). Future work could explore methods to prioritize the development and diversification of more promising skills, potentially leading to a more efficient and practical skill repertoire.

\bibliography{references}

\newpage
\appendix
\onecolumn

\section{Supplementary Analysis}

\begin{figure}
    \centering
    \includegraphics[width=0.75\linewidth]{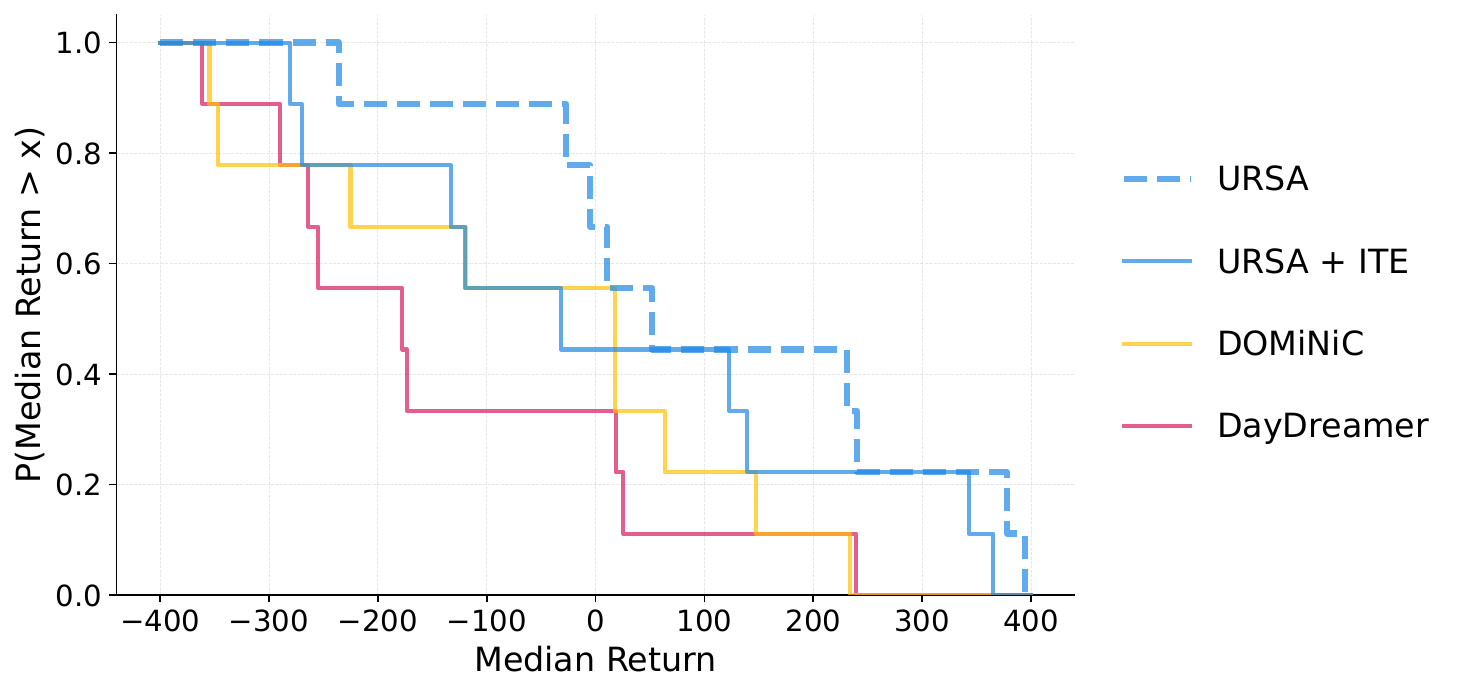}
    \caption{Complementary cumulative distribution functions (CCDF) of the median performance across the simulated damage scenarios from Figure~\ref{fig:damages}.}
    \label{fig:ccdf}
\end{figure}

\ours does not consistently outperform all baselines across individual damage scenarios (Fig.~\ref{fig:damages}). 
Nevertheless, \ours and \oursITE exhibit less performance degradation overall, indicating greater robustness. This is illustrated by the complementary cumulative distribution functions (CCDF) of the median performance across our 9 damage scenarios considered (Fig.~\ref{fig:ccdf}).

The seemingly weaker results from Figure~\ref{fig:damages} stem partly from \ours's network architecture. As \ours uses a skill-conditioned policy network, some discovered skills naturally share similar gaits that rely on specific limbs, making them more vulnerable to targeted damage scenarios. 

\section{Method Details}
\label{appendix:method-details}

\begin{figure*}[t]
    \centering
    \includegraphics[width=\textwidth]{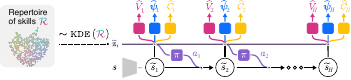}
    \caption{Illustration of URSA’s imagination-based training loop in the world model $\wm$. Given a sampled skill $\skill$, the world model generates imagined trajectories used to update networks parameterizing the value function $\valueFunction$, successor features $\successorFeatures$, cost function $\costCritic$, and the policy $\policy$.}
    \label{fig:ursa_training_actor}
\end{figure*}

Here we provide further details on our optimization objective (Problem~\ref{eq:problem-3}), skill sampling strategy (Section~\ref{sec:method-skill_sampling}), and adaptive thresholding mechanism (Section~\ref{sec:method-skill_sampling}). 

\subsection{Lagrangian Optimization for Safe and Targeted Skill Execution}
\label{appendix:lagrange-details}

We solve the constrained optimization problem in Eq.~\ref{eq:problem-3} using min-max Lagrangian optimization:
\begin{equation*}
    \max_{\policy} \min_{\lagrange_1, \lagrange_2 \geq 0} 
    {\color{perf} \valueFunction(\state, \skill) }
    -  \lagrange_1 ({\color{dist} \norm{(1 - \discount) \successorFeatures(\state, \skill) - \skill}} - \threshold)
    - \lagrange_2 {\color{safe} \costCritic(\state, \skill)}
\end{equation*}
where $\lagrange_1$ and $\lagrange_2$ are the Lagrange multipliers associated with the distance to skill and safety constraints, respectively. 

In practice, we found that optimizing the following smoothed actor objective with clipped multipliers $0 \leq \lagrange_1, \lagrange_2 \leq 1$ improves stability:
\begin{equation*}
    J_\policy = (1 - \lagrange_1)(1 - \lagrange_2) {\color{perf} \valueFunction(\state, \skill) }
    -  \lagrange_1(1 - \lagrange_2)({\color{dist} \norm{(1 - \discount) \successorFeatures(\state, \skill) - \skill}} - \threshold)
    - \lagrange_2 {\color{safe} \costCritic(\state, \skill)}
\end{equation*}

The Lagrange multipliers $\lagrange_1$ and $\lagrange_2$ are parameterized as neural networks that take $(\state, \skill)$ as input. This allows dynamic trade-offs across states and skills. As $\lagrange_2$ increases, the agent prioritizes safety, suppressing both reward maximization and skill execution. Once safety constraints are reliably satisfied (i.e., $\lagrange_2$ is low), the agent shifts focus to skill tracking by increasing $\lagrange_1$. When both constraints are met, the agent emphasizes reward maximization.  During training, they are updated via gradient descent.

\subsection{Kernel Density Estimation Sampling}
\label{appendix:kde_sampling}

\paragraph{Initialization} Initially, the repertoire $\mathcal{R}$ is empty, so we sample skills from a standard normal distribution $\mathcal{N}(\boldsymbol{0}, \boldsymbol{I})$. These initial random skills do not compromise learning since sampled skills are only used for action selection and policy updates, and are never stored in the dataset $\mathcal{D}$. Once the repertoire contains enough skills for a non-degenerate covariance matrix, we transition to sampling from the KDE fitted on $\mathcal{R}$.

\paragraph{KDE Sampling} To sample a skill $\skill_{\text{target}}$ from the KDE distribution $\surrogateProb = \kde{\repertoire} = \frac{1}{\sizeRepertoire} \sum_{i=1}^{\sizeRepertoire} \normal(\skill_i, \mat{\Sigma})$, we use the standard mixture sampling procedure:
\begin{enumerate}
   \item \textbf{Component Selection:} Uniformly select an index $i \sim \text{Uniform}(1, \ldots, \sizeRepertoire)$ from the repertoire.
   \item \textbf{Gaussian Sampling:} Sample $\skill_{\text{target}} \sim \normal(\skill_i, \mat{\Sigma})$ using the selected repertoire skill $\skill_i$ as the mean.
\end{enumerate}

The covariance matrix $\mat{\Sigma}$ is the empirical covariance of all skills in the repertoire scaled by Scott's rule bandwidth $h = \sizeRepertoire^{-\frac{1}{\dimSkillSpace+4}}$, such that the final covariance is $\mat \Sigma = h^2 \mat{\Sigma}_{\text{empirical}}$. This sampling procedure naturally concentrates samples around regions of high skill density while allowing exploration of nearby skill variations through the Gaussian kernels.

\subsection{Repertoire Update}

\begin{algorithm}[t]
\caption{\ours~-- Repertoire Update}
\label{algo:repertoire_update}
\renewcommand{\algorithmiccomment}[1]{\hfill$\triangleright$#1}
\newcommand{\LINECOMMENT}[1]{$\triangleright$#1}

\begin{algorithmic}
\INPUT{state $\state_t$, current repertoire $\repertoire$, maximum size $\sizeRepertoire$}
\OUTPUT{Updated repertoire $\repertoire$}
\item[]
\STATE{$\feat_t = \feat(\state_t)$}
\item[]
\IF{$\feat_t\notin \stateSpaceSafe$}
\STATE{\textbf{return}} \COMMENT{ Never add unsafe features to skill repertoire}
\ELSIF{$|\repertoire| < \sizeRepertoire$}
    \STATE{$\repertoire \leftarrow \repertoire \cup \{\feat_t\}$} \COMMENT{ Add to repertoire if it is not full yet}
\ELSE
    \STATE{$\repertoire' \leftarrow \repertoire \cup \{\feat_t\}$} \COMMENT{ Create temporary extended repertoire}
    \FOR{each skill $\skill \in \repertoire'$}
        \STATE{$d_{\skill} \leftarrow \min_{\skill' \in \repertoire' \setminus \{\skill\}} \distanceFunc_{\mat{\Sigma}}(\skill, \skill')$} \COMMENT{ Compute nearest neighbor distance for each skill}
    \ENDFOR
    \STATE{$\skill_{\text{remove}} \leftarrow \argmin_{\skill \in \repertoire'} d_{\skill}$} \COMMENT{ Find skill with smallest nearest neighbor distance}
    \STATE{$\repertoire \leftarrow \repertoire' \setminus \{\skill_{\text{remove}}\}$} \COMMENT{ Remove to improve diversity and maintain $\sizeRepertoire$ elements}
\ENDIF
\end{algorithmic}
\end{algorithm}

Algorithm~\ref{algo:repertoire_update} details the repertoire update mechanism from Section~\ref{sec:method-skill_sampling}.

\subsection{Imagination-Based Skill Learning}
\label{appendix:imagination-training}

Figure~\ref{fig:ursa_training_actor} illustrates the imagination-based training loop. The world model generates imagined rollouts conditioned on a sampled skill $\skill$, from an initial state $\state$, over a fixed horizon $\horizonImg$. These trajectories are used to train the value function $\valueFunction$, successor features $\successorFeatures$, cost critic $\costCritic$, and policy $\policy$, enabling efficient skill acquisition.

\subsection{Adaptive Threshold Computation}
\label{appendix:adaptive-threshold}

To avoid satisfying multiple constraints with a single skill, we define an adaptive constraint threshold $\threshold$ that scales with the repertoire’s density and the target number of distinct skills $\numDistinctSkills$:
\begin{equation*}
    \threshold = 
    \left( \frac{\sizeRepertoire}{\numDistinctSkills} \right)^{1 / \dimSkillSpace} 
    \cdot \frac{1}{2 \sizeRepertoire} 
    \sum_{i=1}^{\sizeRepertoire} 
    \distanceFunc_{\mat{\Sigma}}(\skill_i, \skill_i^{\mathrm{nn}})
\end{equation*}
Here, $\skill_i^{\mathrm{nn}}$ is the nearest neighbor of $\skill_i$ under the Mahalanobis distance. This formulation ensures that the learned skills are well-separated in the latent space.

\section{Mathematical Derivations}
\label{appendix:math-derivations}

We provide here the technical details behind the lower bound on the approximate entropy of the KDE: 

\begin{align*}
    \estimatedEntropy(\kde{\repertoire}) 
    &= -\frac{1}{\sizeRepertoire} \sum_{i=1}^{\sizeRepertoire} \log \left( \frac{1}{\sizeRepertoire} \sum_{j=1}^{\sizeRepertoire} \normal(\skill_i | \skill_j, \mat{\Sigma}) \right) \\
    &\geq - \frac{1}{\sizeRepertoire} \sum_{i=1}^{\sizeRepertoire} \log \left( 1 + (\sizeRepertoire - 1) e^{ -\frac{1}{2} \distanceFunc_{\mat{\Sigma}}(\skill_i, \skill_i^{\mathrm{nn}})^2 } \right) + \frac{1}{2} \log \left(\det(\mat{\Sigma}) \right) + \frac{\dimSkillSpace}{2} \log(2\pi) + \log \sizeRepertoire
\end{align*}
where $\distanceFunc_{\mat{\Sigma}}(\cdot, \cdot)$ is the Mahalanobis distance with respect to the covariance matrix $\mat{\Sigma}$.

\begin{proof}

    For all $i \in \left\{1, 2, \ldots, \sizeRepertoire\right\}$:
    \begin{align*}
        \log \sum_{j=1}^{\sizeRepertoire} \normal(\skill_i | \skill_j, \mat{\Sigma})
        &= \log \left( \frac{1}{\sqrt{(2\pi)^\dimSkillSpace \det(\mat{\Sigma})}} \sum_{j=1}^{\sizeRepertoire} \exp \left( -\frac{1}{2} (\skill_i - \skill_j)^\intercal \mat{\Sigma}^{-1} (\skill_i - \skill_j) \right) \right) \\
        &= \log \left(\sum_{j=1}^{\sizeRepertoire} \exp \left( -\frac{1}{2} (\skill_i - \skill_j)^\intercal \mat{\Sigma}^{-1} (\skill_i - \skill_j) \right) \right) - \frac{1}{2} \log \left(\det(\mat{\Sigma}) \right) - \frac{\dimSkillSpace}{2} \log(2\pi) \\
        &= \log \left(\sum_{j=1}^{\sizeRepertoire} \exp \left( -\frac{1}{2} \distanceFunc_{\mat{\Sigma}}(\skill_i, \skill_j)^2 \right) \right) - \frac{1}{2} \log \left(\det(\mat{\Sigma}) \right) - \frac{\dimSkillSpace}{2} \log(2\pi) \\
        &= \lse \left( -\frac{1}{2} \distanceFunc_{\mat{\Sigma}}(\skill_i, \skill_1)^2, \ldots, -\frac{1}{2} \distanceFunc_{\mat{\Sigma}}(\skill_i, \skill_{\sizeRepertoire})^2 \right) - \frac{1}{2} \log \left(\det(\mat{\Sigma}) \right) - \frac{\dimSkillSpace}{2} \log(2\pi) \\
        &\leq  \log \left( 1 + (\sizeRepertoire - 1) \exp \left( -\frac{1}{2} \distanceFunc_{\mat{\Sigma}}(\skill_i, \skill_i^{\mathrm{nn}})^2 \right) \right) - \frac{1}{2} \log \left(\det(\mat{\Sigma}) \right) - \frac{\dimSkillSpace}{2} \log(2\pi) \\
    \end{align*}

    Then we have:
    \begin{align*}
        \estimatedEntropy(\kde{\repertoire}) 
        &= -\frac{1}{\sizeRepertoire} \sum_{i=1}^{\sizeRepertoire} \left( \log \sum_{j=1}^{\sizeRepertoire} \normal(\skill_i | \skill_j, \mat{\Sigma}) \right) + \log \sizeRepertoire \\
        &\geq - \frac{1}{\sizeRepertoire} \sum_{i=1}^{\sizeRepertoire} \log \left( 1 + (\sizeRepertoire - 1) e^{ -\frac{1}{2} \distanceFunc_{\mat{\Sigma}}(\skill_i, \skill_i^{\mathrm{nn}})^2 } \right) + \frac{1}{2} \log \left(\det(\mat{\Sigma}) \right) + \frac{\dimSkillSpace}{2} \log(2\pi) + \log \sizeRepertoire
    \end{align*}

\end{proof}

\section{Training Details}
\label{appendix:training-details}

This section provides implementation details for the reward, feature, and cost functions used in both unsupervised and velocity-conditioned experiments, as referenced in Section~\ref{sec:results}. We also report details on the \oursITE implementation used for damage adaptation tasks in Section~\ref{sec:results-rq2}. Additional hyperparameters are listed in Table~\ref{tab:hyperparameters-ours}.

\subsection{Unsupervised Quality-Diversity}
\label{appendix:reward-unsupervised}

For the unsupervised setting, our objective is to learn a repertoire of diverse behaviors for forward locomotion. The feature function $\featFunc(\cdot)$ is implemented as a VAE encoder network that maps the robot’s joint angles and torso height at each timestep $t$ into a 2D latent skill space $\skillSpace$.

The reward function promotes forward movement while ensuring stability:
\begin{equation*}
    r(s_t, a_t) = r_{\text{upr}} + \mathbbm{1}_{r_{\text{upr}} > 0.7} \cdot (5r_{\text{vel}_x} - 0.5r_{\text{vel}_y} - 0.5r_{\text{yaw}}) - 0.001(r_{\text{speed}} + r_{\text{work}} + r_{\text{smooth}})
\end{equation*}
Here, $r_{\text{upr}}$ encourages upright posture, while forward velocity $r_{\text{vel}_x}$ is rewarded. Lateral and yaw motion ($r_{\text{vel}_y}, r_{\text{yaw}}$) are penalized to promote forward-aligned behavior. Smoothness penalties ($r_{\text{speed}}, r_{\text{work}}, r_{\text{smooth}}$) encourage efficient movement. Following~\citet{daydreamer}, velocity-based rewards are only active when the robot is upright ($r_{\text{upr}} > 0.7$).

The safe state set includes all states with $r_{\text{upr}} > 0.7$, and the cost function is defined as:
\[
    c(s_t, a_t) = 0.7 - r_{\text{upr}}(s_t, a_t)
\]
This penalizes instability and unsafe motion when the robot deviates from an upright posture. We aim to learn a repertoire of $\numDistinctSkills = 512$ distinct skills in this setting.

\subsection{ITE for Damage Adaptation}
\label{appendix:ite}

In this section, we provide the implementation details of the iterative trial and error (ITE) algorithm, adapted for use with \ours's skill repertoire. The approach is based on the methodology introduced in \citet{cully_RobotsThatCan_2015}.

The ITE process is run for 8 iterations to match the number of skills in the \dominic baseline. We set the length scale parameter to $0.1$ in the Gaussian process model to control the smoothness of the predictions and the noise level to $1 \times 10^{-3}$ to account for observation noise during optimization.

We build a behavior map that serves as our prior, constructed by evaluating value function returns across 512 rollouts in our repertoire. This matches the number of distinct skills used during training. 

For the kernel used in the Gaussian process, we employ a Matern52 kernel, which has proven effective in many real-world scenarios due to its flexibility and smoothness properties. In terms of the acquisition function, we adopt the Upper Confidence Bound (UCB) approach, following the original implementation in~\citep{cully_RobotsThatCan_2015}. This acquisition function balances exploration and exploitation by selecting the point with the highest expected return, considering both the predicted mean and uncertainty of the model.

These steps ensure that the ITE algorithm efficiently adapts to damage while leveraging the learned skill repertoire for robust performance in real-world environments.

\subsection{Velocity-Conditioned Skill Discovery}
\label{appendix:reward-velocity}

For the heuristic-based skill discovery setting, the reward function encourages smooth and well-postured motion:
\begin{equation*}
    r(s_t, a_t) = r_{\text{upr}} + r_{\text{hip}} + r_{\text{upper}} + r_{\text{lower}} - 0.001(r_{\text{speed}} + r_{\text{work}} + r_{\text{smooth}})
\end{equation*}
Each term encourages proper posture at a different joint level (torso, hip, upper leg, lower leg), activated sequentially through dependency on earlier posture terms (e.g., $r_{\text{hip}}$ activates only if $r_{\text{upr}} > 0.7$). This encourages the robot to build stable poses from the ground up.

Safe states are defined by the conditions: 
\[
r_{\text{upr}},\ r_{\text{hip}},\ r_{\text{upper}},\ r_{\text{lower}} > 0.7
\]
The cost function penalizes deviations below this threshold:
\[
    c(s_t, a_t) = \sum_{j \in \{\text{upr}, \text{hip}, \text{upper}, \text{lower}\}} \max(0.7 - r_j,\ 0)
\]

\subsection{Hyperparameters}
\label{appendix:hyperparameters}

A full list of rollout parameters and training constants used in our experiments is provided in Table~\ref{tab:hyperparameters-ours}.

\begin{table}[H]
\caption{Hyperparameters}
\label{tab:hyperparameters-ours}
\centering
\begin{tabular}{l | c}
\toprule
Parameter & Value\\
\midrule
Imagination batch size $N$ & $1024$\\
Real environment exploration batch size & $32$\\
Total timesteps (in simulation) & $1 \times 10^{6}$ \\
Optimizer & Adam\\
Learning rate & $1 \times 10^{-4}$\\
Replay buffer size & $1 \times 10^6$\\
Discount factor $\discount$ & $0.995$\\
Imagination horizon $H$ & $15$\\
Target smoothing coefficient $\tau$ & $0.02$\\
Sampling period $T$ & $250$\\
Lambda Return $\lambda$ & $0.95$\\
Repertoire size $\sizeRepertoire$ & 4096 \\
Target distinct skills $\numDistinctSkills$ & 512 \\
Control frequency & 20 Hz \\
\midrule
\multicolumn{2}{c}{\dominic \citep{dominic}} \\
Number of skills $n_{\mathrm{dominic}}$ & 8 \\
Near-optimality constraint & $\geq 0.9 \times \text{value optimal policy}$ \\
Target distance between skills $l_0$ & 1.0 \\
\bottomrule
\end{tabular}
\end{table}


\end{document}